\documentclass[letterpaper, 10 pt, conference]{ieeeconf}
\IEEEoverridecommandlockouts
\title{\LARGE \bf Enhancing Autonomous Navigation by Imaging Hidden Objects \\ using Single-Photon LiDAR}

\author{Aaron Young$^{*}$$^{1}$, Nevindu M. Batagoda$^{*}$$^{2}$, Harry Zhang$^{2}$, Akshat Dave$^{1}$, \\ Adithya Pediredla$^{3}$, Dan Negrut$^{2}$, Ramesh Raskar$^{1}$%
\thanks{$^{*}$These authors contributed equally.}
\thanks{$^{1}$MIT Media Lab, Massachusetts Institute of Technology, 75 Amherst St, Cambridge, MA 02139 {\tt\small {aryoung, ad74, raskar}@mit.edu}}
\thanks{$^{2}$Department of Mechanical Engineering, University of Wisconsin - Madison, 1513 University Ave, Madison, WI 53706. {\tt\small {batagoda, hzhang699, negrut}@wisc.edu}}
\thanks{$^{3}$Department of Computer Science, Dartmouth College, 15 Thayer Dr, Hanover, NH 03755. {\tt\small adithya.k.pediredla@dartmouth.edu}}
}

\let\labelindent\relax

\usepackage{amsmath}
\usepackage{xcolor}
\usepackage{graphicx}
\usepackage{cleveref}
\usepackage{paralist}
\usepackage{enumitem}
\usepackage{url}

\usepackage{tikz}

\newcommand\submittedtext{%
  \footnotesize This work has been submitted to the IEEE for possible publication. Copyright may be transferred without notice, after which this version may no longer be accessible.}

\newcommand\submittednotice{%
\begin{tikzpicture}[remember picture,overlay]
\node[anchor=south,yshift=10pt] at (current page.south) {\fbox{\parbox{\dimexpr0.65\textwidth-\fboxsep-\fboxrule\relax}{\submittedtext}}};
\end{tikzpicture}%
}

\newif\ifshowcomments
\showcommentstrue 
\showcommentsfalse

\ifshowcomments
    \newcommand{\comment}[1]{\textcolor{olive}{{\em #1}}}
    \newenvironment{multilinecomment}[1]{\begingroup\color{olive}#1}{\endgroup}
    \newcommand{\AY}[1]{\textcolor{blue}{{\em {\bf AY:} #1}}}
    \newcommand{\AD}[1]{\textcolor{purple}{{\em {\bf AD:} #1}}}
    \newcommand{\AP}[1]{\textcolor{orange}{{\em {\bf AP:} #1}}}
    \newcommand{\NB}[1]{\textcolor{pink}{{\em {\bf NB:} #1}}}
    \newcommand{\HZ}[1]{\textcolor{RubineRed}{{\em {\bf HZ:} #1}}}
    
    \newcommand{\oneline}[1]{{\textcolor{Emerald}{\textbf{One-Liner:} {#1}}}}
\else
    \newcommand{\comment}[1]{}
    
    \newcommand{\AY}[1]{}
    \newcommand{\AD}[1]{}
    \newcommand{\AP}[1]{}
    \newcommand{\NB}[1]{}
    \newcommand{\HZ}[1]{}
    
    \newcommand{\oneline}[1]{}
\fi

\begin{document}

\maketitle
\submittednotice
\thispagestyle{empty}
\pagestyle{empty}


\begin{abstract}
Robust autonomous navigation in environments with limited visibility remains a critical challenge in robotics. We present a novel approach that leverages Non-Line-of-Sight (NLOS) sensing using single-photon LiDAR to improve visibility and enhance autonomous navigation. Our method enables mobile robots to ``see around corners" by utilizing multi-bounce light information, effectively expanding their perceptual range without additional infrastructure. We propose a three-module pipeline: (1) Sensing, which captures multi-bounce histograms using SPAD-based LiDAR; (2) Perception, which estimates occupancy maps of hidden regions from these histograms using a convolutional neural network; and (3) Control, which allows a robot to follow safe paths based on the estimated occupancy. We evaluate our approach through simulations and real-world experiments on a mobile robot navigating an L-shaped corridor with hidden obstacles. Our work represents the first experimental demonstration of NLOS imaging for autonomous navigation, paving the way for safer and more efficient robotic systems operating in complex environments. We also contribute a novel dynamics-integrated transient rendering framework for simulating NLOS scenarios, facilitating future research in this domain.
\end{abstract}


\section{Introduction}
\label{sec:introduction}


\begin{figure*}
    \centering
    \includegraphics[width=\linewidth]{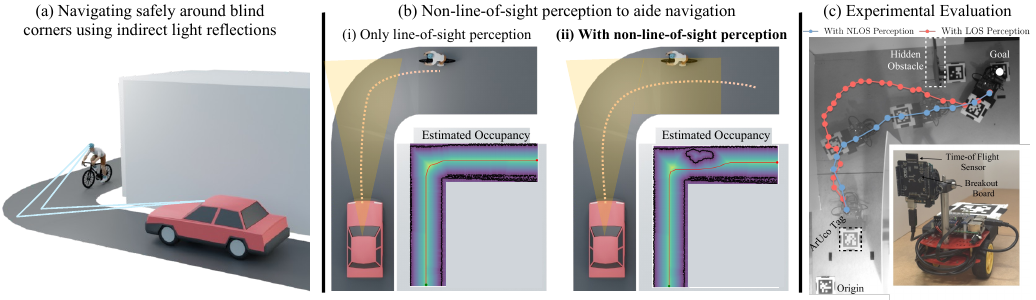}
    \caption{\textbf{Autonomous navigation using single-photon LiDAR.} (a) Autonomous navigation is challenging under unseen threats such as a cyclist emerging from a blind corner. We leverage single-photon LiDAR for autonomous navigation by seeing around corners using indirect reflections of light. (b) With only line-of-sight perception, the location of hidden objects around the corner is unknown to the agent, resulting in collisions. Our approach enables non-line-of-sight (NLOS) perception with the estimated occupancy map accurately estimating the hidden object location. (c) We evaluate our NLOS-based navigation approach experimentally using a commercial-grade single-photon LiDAR mounted on a two-wheeled robot. Incorporating NLOS perception results in a smoother trajectory while avoiding the hidden obstacle.}
    \label{fig:teaser}
\end{figure*}

Robust autonomous navigation remains a critical challenge in robotics as mobile robots become increasingly prevalent, from industrial warehouses to urban environments. These systems have the potential to save lives by preventing accidents, reduce financial losses by protecting human and robotic assets, and enhance operational efficiency by optimizing operation speed and energy consumption. 


A fundamental challenge in achieving robust autonomous navigation lies in the inherent limits of perception systems. A robot's actions are ultimately constrained by what it can perceive. When unseen threats or obstacles suddenly appear in the robot's path, safe maneuvers become challenging. For example, consider a pedestrian emerging from a blind corner in front of a self-driving car. The autonomy stack must be capable of rapid detection and response, despite the threat being previously outside its field of view. The multi-bounce LiDAR approach outlined in this paper aims to address such challenges by expanding the robot's perceptual capabilities, potentially allowing it to ``see around corners" and react to threats before they become imminent dangers.

Existing solutions for autonomous navigation under limited visibility come with trade-offs. One common approach is to simply reduce speed at blind corners or in areas with potential hidden hazards. While this strategy improves safety, it comes at the cost of reduced efficiency, wasting both time and energy. Another solution is to use Vehicle-to-Everything (V2X) communication, which can provide additional information about the environment beyond the robot's immediate field of view. However, such solutions often require the installation and maintenance of additional infrastructure, which can be costly and may not be feasible in all environments. 


We propose a novel solution that addresses the limitations of both speed reduction and additional infrastructure requirements. Drawing inspiration from the concept of safety mirrors used at blind intersections, we leverage Non-Line-of-Sight (NLOS) sensing to enhance autonomous navigation capabilities. The key insight of our work is that multi-bounce LiDAR technology can effectively transform ordinary walls and surfaces into virtual safety mirrors, expanding the robot's perceptual range without the need for additional infrastructure. 

NLOS sensing utilizes the reflections of light from surfaces to detect objects that are not directly visible to the sensor. LiDAR emits laser pulses that bounce off walls and other surfaces before returning to the sensor, carrying information about occluded areas. A LiDAR sensor measures the time-of-flight of the returned light pulses. From the time-of-flight of multi-bounce light, one can reconstruct the presence and movement of objects around corners or behind obstacles.

We propose a novel pipeline that leverages single-photon LiDAR to enable non-line-of-sight perception in autonomous navigation, see \Cref{fig:teaser}. Our approach has three key modules operating in a loop: 1) Sensing, which captures multi-bounce histograms from a SPAD-based LiDAR; 2) Perception, which estimates occupancy maps of hidden regions from these histograms; and 3) Control, which allows an agent to follow safe paths based on the estimated occupancy maps.


We evaluated the effectiveness of our NLOS perception approach for autonomous navigation through simulation and real-world experiments. We develop a simulation framework that couples transient rendering with vehicle dynamics within the multi-physics engine Project Chrono~\cite{chrono,chsensor}. This allows us to co-simulate and model the multi-bounce histograms of single-photon LiDARs with vehicle/robot dynamics simulations. Our simulation experiments demonstrate NLOS perception aids in navigating an L-turn corner with reduced chances of collision with hidden obstacles compared to LOS-only perception. We implemented our system on a physical two-wheeled robot equipped with a commercial-grade single-photon LiDAR. In a blind corner navigation task, the NLOS-capable robot demonstrated superior performance, generating and following more efficient paths around hidden objects. Quantitatively, the NLOS approach reduced travel time by more than half and shortened the trajectory by 33\% compared to LOS-only navigation.  

Our contributions are summarized as follows:
\begin{itemize}[leftmargin=10pt]
    \item Ours is the first work to explore the role of non-line-of-sight perception for autonomous navigation.
    \item We propose a novel pipeline for incorporating multi-bounce histograms containing NLOS information into a navigation framework, including a data-driven NLOS perception approach. 
    \item We develop a transient rendering engine within the open-source multi-physics engine Project Chrono to simulate single-photon LiDARs and robot dynamics in conjunction. Through simulations, we demonstrate our approach reduces the likelihood of collisions compared to LOS-only perception.  
    \item We experimentally evaluate our NLOS-aided approach on a novel setup leveraging commercial-grade single-photon LiDAR mounted on a two-wheeled robot.
\end{itemize}




\section{Related Work}
\label{sec:background}

\subsection{Single-photon LiDAR}
Single-photon LiDAR has emerged as a promising sensing technology for robotics applications. Unlike traditional LiDAR systems that typically record only the first return, single-photon LiDARs can detect and time-stamp individual photons with picosecond-level precision~\cite{rapp2020advances, callenberg2021low}. Single-photon LiDAR has higher sensitivity and allows one to capture not just the direct reflections from visible surfaces, but also the indirect reflections that have undergone multiple bounces before returning to the sensor. Single-photon LiDAR has shown great promise in depth sensing~\cite{lindell2018single, rapp2021high}, 3D surface mapping~\cite{Sifferman2024using, Mu24Towards3DVision, behari2024blurred}, handling specular surfaces~\cite{henley2023detection, lin2024handheld}, long-range detection~\cite{hadfield2023single, chan2019long}, airborne applications such as snow depth mapping~\cite{king2022evaluation, king2023new, henley2023detection}, forest surveying~\cite{swatantran2016rapid}, and 3D terrain mapping~\cite{hong2024airborne}.

Our system specifically utilizes \textit{flash} single-photon LiDARs that are now commercially available and inexpensive \cite{ams_osram_tmf882x, st_vl53l4cx}. These LiDARs feature flash illumination that floods the scene, coupled with an array of SPAD pixels. This configuration enables simultaneous capture of spatio-temporal multi-bounce information. 

\subsection{Non-Line-Of-Sight Imaging}
Non-line-of-sight imaging has emerged as a promising application by utilizing single-photon LiDAR that can measure indirect reflections and image objects hidden from the direct line of sight. Recent works demonstrate 3D reconstruction of occluded scenes from single-photon LiDAR at high spatial resolution and computational efficiency \cite{o2018confocal,heide2019non,lindell2019wave, liu2019non, pediredla2019snlos, ahn2019convolutional, PlatoNeRF, somasundaram2023role,metzler2020keyhole,velten2012recovering,liu2020phasor}. While the majority of existing works focus on 3D imaging, there are also notable contributions to 3D pose estimation~\cite{isogawa2020optical, liu2022hiddenpose} and 3D tracking~\cite{callenberg2021low} using NLOS information. However, these works typically rely on expensive research-grade sensors. Only a few works have shown NLOS imaging with commercial-grade sensors~\cite{callenberg2021low}. Our paper presents the first experimental demonstration of NLOS imaging for robotics applications, paving the way for novel applications in autonomous navigation and sensing.
  
\begin{figure*}[!ht]
    \centering
    \includegraphics[width=\linewidth]{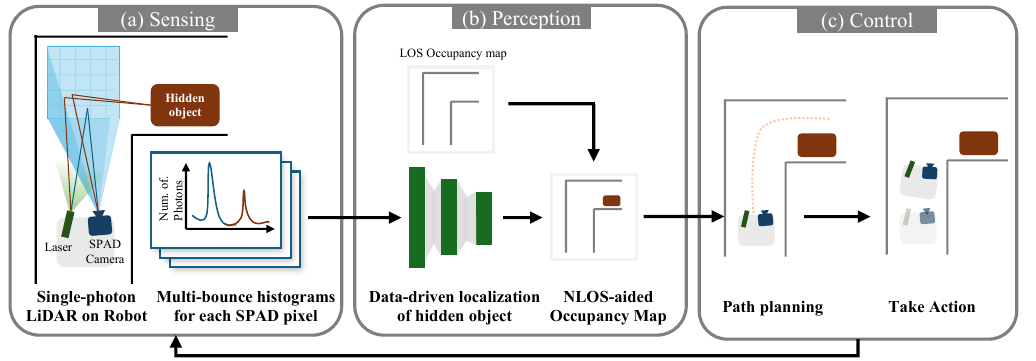}
    \caption{\textbf{Our pipeline for NLOS-aided Autonomous Navigation.} (a) Sensing: We capture multi-bounce histograms using single-photon LiDAR to gather information about hidden regions. (b) Perception: Captured histograms are then processed using a data-driven approach to estimate the occupancy map of the occluded region. (c) Control: Then the optimal path is planned based on the estimated occupancy map for navigating around obstacles.}
    \label{fig:pipeline}
\end{figure*}

\subsection{Safety-Critical Control}

Developing autonomous systems capable of collision-free obstacle avoidance is challenging and remains an area of on-going research. One approach is to integrate vehicle path tracking, stabilization, and collision avoidance, temporarily prioritizing collision avoidance over stabilization~\cite{funkecollisionavoidance2017}. Model predictive control (MPC) is a proven technique which has been shown to be an effective controller for vehicles at high speeds~\cite{kim2023trajectory,micheli2023nmpc}, in highway scenarios~\cite{li2023research}, and for collision avoidance~\cite{quan2021robust,grandia2021multi}. While MPC-based algorithms couple planning and tracking, treating them as separate problems has proven simple and reliable. Common planning algorithms employed in mobile robotics include Dijkstra's algorithm, A*, and the Artificial Potential Field (APF) method~\cite{rafai2022review,zhang2023global,zhang2024agv,ji2023tripfield,kabir2023enhanced}, amongst others. While many control-focused strategies aim to improve obstacle avoidance through control design, e.g, predictive control and control barrier functions, we approach this problem by enhancing perception to increase visibility, giving control algorithms more reaction time and space.


\section{NLOS-aided Autonomous Navigation}
\label{sec:methods}

In this section, we present our pipeline for incorporating non-line-of-sight perception into autonomous navigation, see also Fig.~\ref{fig:pipeline}. We utilize single-photon LiDAR to enhance autonomous agent's perception beyond its line of sight. Our pipeline consists of three key modules that operate in a continuous loop: Sensing, Perception, and Control. In the \textit{Sensing} phase, the agent captures multi-bounce histograms from a SPAD-based LiDAR, gathering information about the hidden region. Subsequently, the \textit{Perception} module uses these histograms to estimate an occupancy map of the occluded areas. Finally, the \textit{Control} module plans a path based on the estimated occupancy. This process repeats at each time step, allowing the agent to effectively plan for and navigate around obstacles in blind corners. The following subsections describe each of the modules in detail.

\subsection{Sensing histograms using single-photon LiDAR}
For the single-photon LiDARs, the multi-bounce information is captured in the form of histograms, which record the number of photon arrivals at different time bins. These histograms encode both the direct and indirect light transport. In a typical histogram, the first peak corresponds to the direct component -- light that has reflected directly off visible surfaces. Subsequent peaks typically represent light that has bounced three times before reaching the sensor: \begin{inparaenum}[1)]
\item from the light source to the visible surface;
\item from the visible surface to the hidden surface; and 
\item from the hidden surface to the visible surface before reaching the LiDAR detector.
\end{inparaenum}
The position of the three-bounce peaks in the histogram encodes location and size of the hidden or occluded objects.

We utilize \textit{Flash} single-photon LiDARs  \cite{ams_osram_tmf882x, st_vl53l4cx}, which comprises of an array of SPAD pixels that each generates its own time-of-flight histogram. In this paper, we consider that the SPAD LiDAR is pointed at a planar floor, and that there is single hidden object of unknown location $(x_h, y_h)$. The spatio-temporal histogram array obtained $\mathbf{H}$ thus depends on $x_h$ and $y_h$.

\begin{align}
    \left(\hat{x}_h, \hat{y}_h\right) = f_\theta\left(\mathbf{H}\left(x_h, y_h\right)\right)
\end{align}

\subsection{Perception of the hidden region}
Estimating location of a hidden object from the histogram array is challenging. The first challenge is tied to the ambiguity of the signal signature. There could be multiple hidden object positions that have the same 3-bounce time-of-flight. The second challenge is that of low signal quality. Commercial LiDARs operate with eye-safe lasers imposing a limit on the emitted laser power. Three bounces on surfaces severely attenuates the signal reaching back to the sensor resulting in very low signal-to-noise ratio (SNR).

We tackle the ambiguity and SNR challenges of commercial-grade LiDAR sensing using a data-driven approach. We train a convolutional neural network (CNN) that inputs the histogram array and outputs the location of the hidden object. This network is trained on a dataset of histogram arrays captured by a LiDAR with at different positions with varying hidden object locations and sizes.

The neural network is comprised of two convolutional layers, followed by two fully connected layers. The fully connected layers take the output of the convolutions concatenated with the ground truth position of the robot at the current state. We use L2 loss with respect to the object's ground truth position and orientation.


\subsection{Control based on the estimated occupancy}
\label{sec:control}

We decompose the control task into two stages: path planning and actuation. This two-stage approach allows us to build a controller agnostic to the underlying vehicle in use. We use a simple global path planning algorithm (specifically A* \cite{kabir2023enhanced}), and iteratively optimize the path as the agent moves within the environment. The path produced by A* doesn't take vehicle dynamics under consideration, so we smooth the output path using a Bezier curve. To determine the next actuation signal sent to the vehicle, we sample the first control point and compute the heading error. Based on a proportional feed-forward calculation, we compute a throttle and steering command which is sent to the vehicle.

\section{Simulation Results}
\label{sec:results}


\begin{figure}
    \centering
    \includegraphics[width=\linewidth]{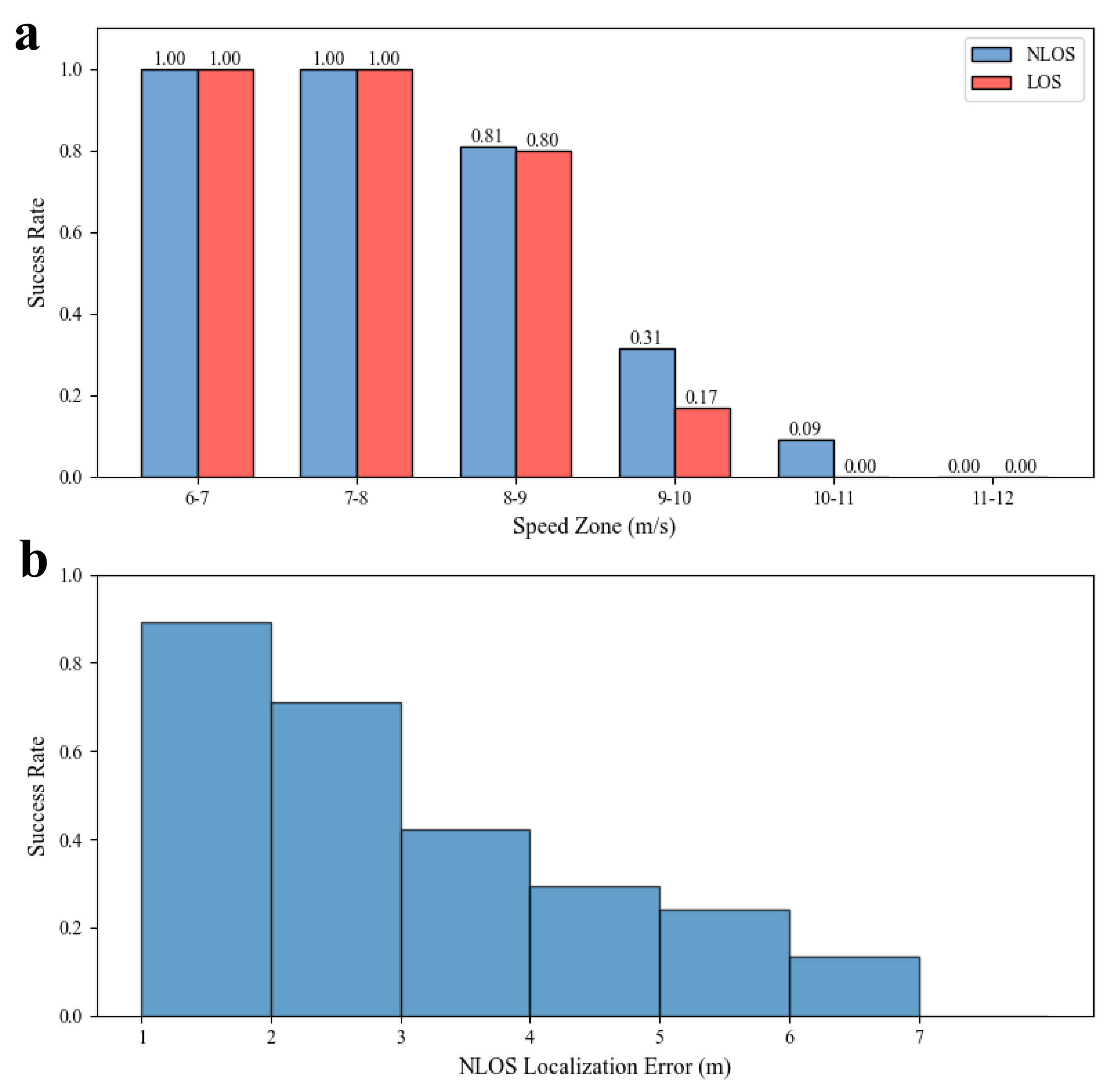}
    \caption{\textbf{Simulation Results for navigating a L-turn corner using line-of-sight and non-line-of-sight perception.} (a) Comparison of collision avoidance success rate between LOS and NLOS perception. NLOS has an edge in performance over LOS at higher speeds where path planning cannot react on time. (b) And, we find that higher localization error leads to worse performance in avoiding obstacles. Which matches intuition.}
    \label{fig:results}
\end{figure}


In this section, we describe our simulation environment, which was developed to test the integration of multi-bounce sensor information for robotic navigation. To accomplish this task, we require two simulators: 1) a dynamics engine capable of simulating a virtual world and virtual agents, and 2) a rendering engine that can model transient data, i.e., the time-of-flight of light within the virtual world.

\subsection{Dynamics-Integrated Transient Rendering}

To design robotic controllers that utilize transient data in-the-loop, we have developed a transient renderer within the open-source Chrono multi-physics engine \cite{chrono}. With native integration of other sensors \cite{chsensor}, vehicles \cite{chvehicle}, and ROS \cite{ros,elmquist2022ros}, Chrono enables the simulation of a wide range of robotic applications. Transient rendering is the simulation of the propagation of light at very high time resolutions. As opposed to standard path tracing, transient rendering models the light at a point in the spatial and time domains ($x,y,t$). Using transient rendering, we can model the amount of light arriving at a time-of-flight detector often at time resolutions on the order of picoseconds or less and record it as a collection of histograms for each time interval. Using these high-time-resolution histograms, we can distinguish light coming from multiple scattering/bounce events in the scene. This enables us to infer information about the scene even beyond the line of sight. Our transient renderer is based on the non-line of sight transient rendering introduced by Royo et al. \cite{royo2022non}. Since NLOS transient rendering is computationally expensive, we use GPU hardware accelerated ray tracing and special sampling methods~\cite{royo2022non} to improve rendering speeds.  Our simulation framework can model vehicle dynamics and render an $8 \times 8$ multi-bounce transient sensor at 10 FPS. 

\subsection{High-speed Blind Corner Navigation with Occluded Obstacles}
We view the use of multi-bounce light as an enhancement to existing perception-based approaches for robotic navigation. Our simulated experiments, shown in \Cref{fig:teaser}, deploy a full-scale wheeled vehicle in a cityscape environment with a randomly placed hidden cyclist wearing retro-reflective gear. We approximate the cyclist with a 3D bounding volume of length 2m. The width and height of the simulation scene is approximately 50m. The vehicle maintains a constant speed throughout the maneuver. The vehicle is then either provided sensor data from a single-bounce LiDAR (\Cref{fig:teaser}a), or from a multi-bounce LiDAR (\Cref{fig:teaser}b). The perceived area is conceptualized as an occupancy map, where obstacles and walls are considered ``occupied'' and everywhere else is ``unoccupied''. Training data for the NLOS object detection model is acquired by running the vehicle using a controller with privileged information on obstacle positions through 30 different environments with unique object positions and saving the transient information. With the trained model, we run a total of 200 trials with uniformly sampled speed values for both LOS and NLOS perception. Analysis of this NLOS vs LOS-based navigation is shown in \Cref{fig:results}a.

In the simulated environment, we found improvement for vehicles capable of perceiving multi-bounce light versus only line-of-sight information. In the $8-9 ms^{-1}$ speed zone, the collision rate decreased by 1\%, by 14\% in the $9-10 ms^{-1}$ zone, and by 9\% in the $10-11 ms^{-1}$ zone as compared to the LOS case. This indicates that being able to detect hidden objects is beneficial when navigating corners while maintaining higher speed profiles. In fact, we see that LOS completely fails to avoid the obstacle at speeds higher than $10 ms^{-1}$. For this particular scenario, the maximum reasonable speed a vehicle could take this corner at was calculated to be $12ms^{-1}$ based on the formula,

\begin{align}
  v_{max} = \sqrt{\mu g r_{max}}
\end{align}

where $r_{max}$ is the maximum turn radius of the corner ($15m$), $\mu$ is the coefficient of friction of the vehicle's tires (estimated to be $0.9$), and $g$ is acceleration due to gravity. As such, for our simulation experiments we set $12ms^{-1}$ as the maximum velocity of the vehicle. 


We examine the effect of localization error on collision rate, where the localization error is the mean of all the errors between ground truth obstacle position and a predicted measurement at a given sensor capture time. When localizing the object position, we define a radius of uncertainty ($r=1m$ in our trials) as the model doesn't output object size. We found that the localization error did not have a considerable effect until the radius of uncertainty was larger than the object itself (\Cref{fig:results}b). As such, we see a degrading performance in obstacle avoidance with high localization errors. We expect simple filtering techniques of the noisy localization values to improve this result.

\section{Experimental Results}

To validate our findings from simulation and assess the viability of multi-bounce LiDAR measurements in robotic navigation, we test our method using a real-world setup. This section describes the results of these experiments, demonstrating the potential of three-bounce measurements for enhancing autonomous navigation systems.

\subsection{Experimental Setup}

Our experimental setup is composed of a two wheeled mobile robot navigation around an L-shaped corridor. We utilize the AMS TMF8828 SPAD-based LiDAR module and their Arduino evaluation kit as the sensor system \cite{ams_osram_tmf882x}. The TMF8828 sensor is comprised of $18 \times 12$ individual SPAD pixels, which is aggregated into $3 \times 3$ measurement zones. In this configuration, the sensor has an effective FoV of $41^\circ \times 52^\circ$ and can measure light traveling a total of 10 m. We acquire sensor information at roughly 5 Hz. Ground truth position measurements are obtained via four ArUco tags \cite{garrido2014aruco} -- placed on the scene, robot, object, and goal respectively -- and imaged with a monochrome camera mounted with a birds-eye view of the entire scene. The robot is a small, two wheeled mobile robot \cite{duckiebot} equipped with a Raspberry Pi 5 for processing. Training of the underlying perception model is performed on an external system, but data acquisition and evaluation takes place on the robot with ground truth positions streamed over a network. Inference of the network on the Raspberry Pi 5 takes approximately 50 milliseconds.


\subsection{Data Acquisition and Training}

\begin{figure}
    \centering
    \includegraphics[width=\linewidth]{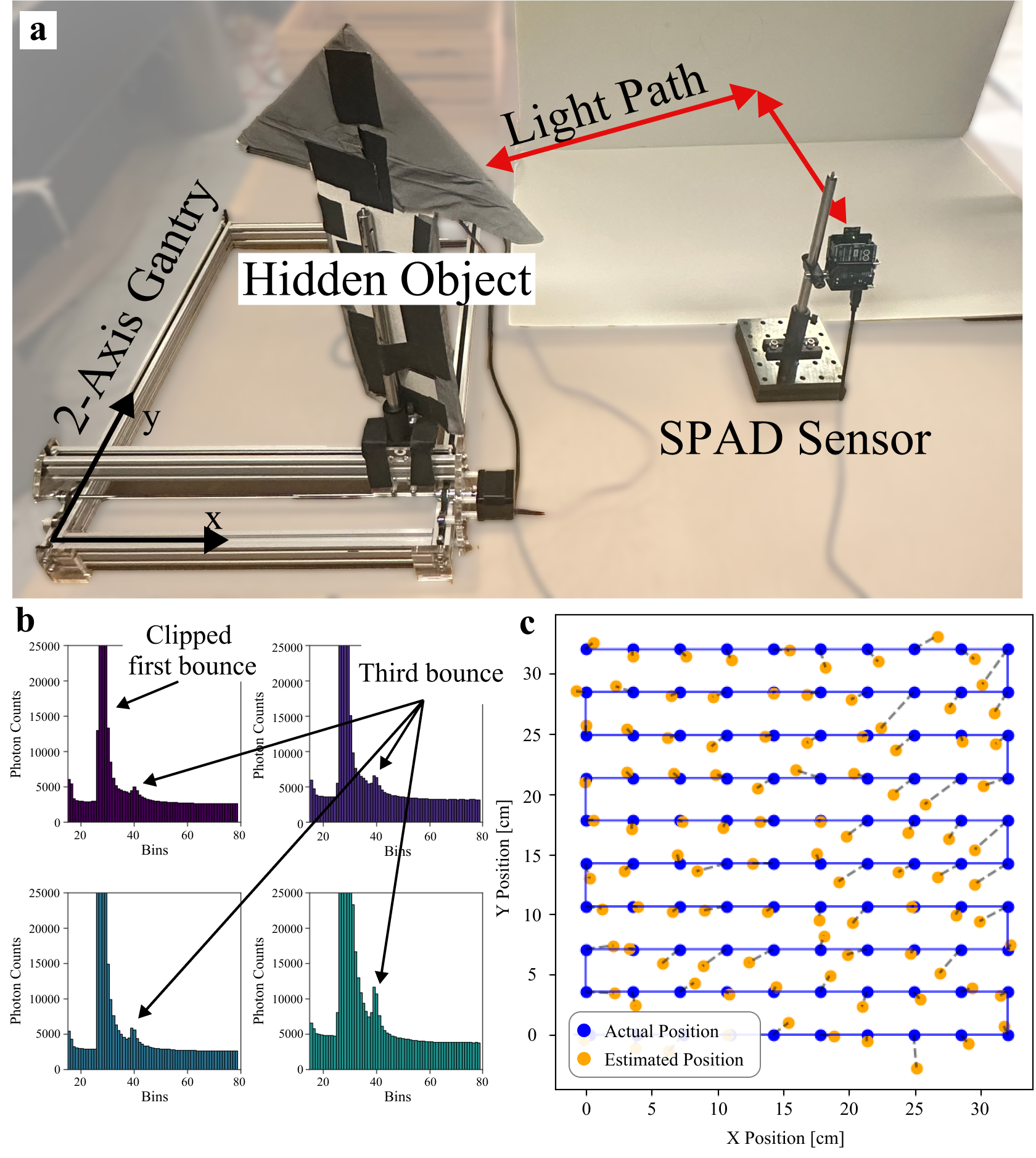}
    \caption{\textbf{Capture and evaluation setup.} (a) To validate the perception algorithm and generate a dataset to train the CNN, we utilize a 2-axis gantry to precisely position a hidden object (in this case, a retro-reflective arrow) in some region outside the FoV of the SPAD sensor. For training, we generate data for many positions and orientations of both the sensor and object to produce a diverse dataset. (b) Multi-bounce histograms of four of the output zones from the SPAD sensor. The first bounce is the primary spike, and in later bins (i.e. with a longer time-of-flight), the third bounce reflection from the object can be observed. Different zones have a different view of the scene, and so the relative intensity and peak position encode the object's state. This information is leveraged by the model to predict the position of the hidden object. (c) The estimates of our model on an evaluation set (i.e., data held out from training) which shows our model estimates object position with sub-10 cm accuracy. We found that evaluation performance decreased substantially when testing on states (i.e., camera/object positions) not seen during training.}
    \label{fig:real-evaluation}
\end{figure}

Data acquisition is performed for 400 unique object positions and 5 unique camera positions within the scene for a total of 2000 samples. A 2-axis gantry system (\Cref{fig:real-evaluation}a) is used to repeatedly position the object, and the mobile robot is used to position the camera during data collection. Each data capture contains 10 exposures which are averaged to obtain the final measurement value for that state. The object itself is covered in a retro-reflective material to enhance third-bounce measurements (\Cref{fig:real-evaluation}b). This setup, using a fixed sensor, is shown in \Cref{fig:real-evaluation}a.

The acquired dataset is subsequently used to train the CNN. The model takes as input the full, uncropped histogram and the ground truth camera position, and is trained to predict the object's x, y, and yaw. During training, the dataset is split 80/20 into a training and evaluation set, respectively. We use a batch size of 16, a learning rate of $1 \times 10^{-3}$, and train for $\sim2000$ epochs. An example evaluation scenario is shown in \Cref{fig:real-evaluation}c, where the model is capable of predicting object states not in it's training dataset to within $5cm$.

We found our method was effective for multiple sensor placements (i.e. height relative to the relay surface) and orientations. During training we use multiple sensor poses, such as oblique views of the ground and perpendicular views of a vertical relay wall. This produced a more robust policy that was effective in interpolating between training locations. However, this required a considerable amount of data. Additionally, object or camera positions out of the bounds of the training dataset produced erroneous predictions. We believe the is due in part to the small size of the CNN in use.

\subsection{Blind Corner Navigation}

\begin{figure}
    \centering
    \includegraphics[width=\linewidth]{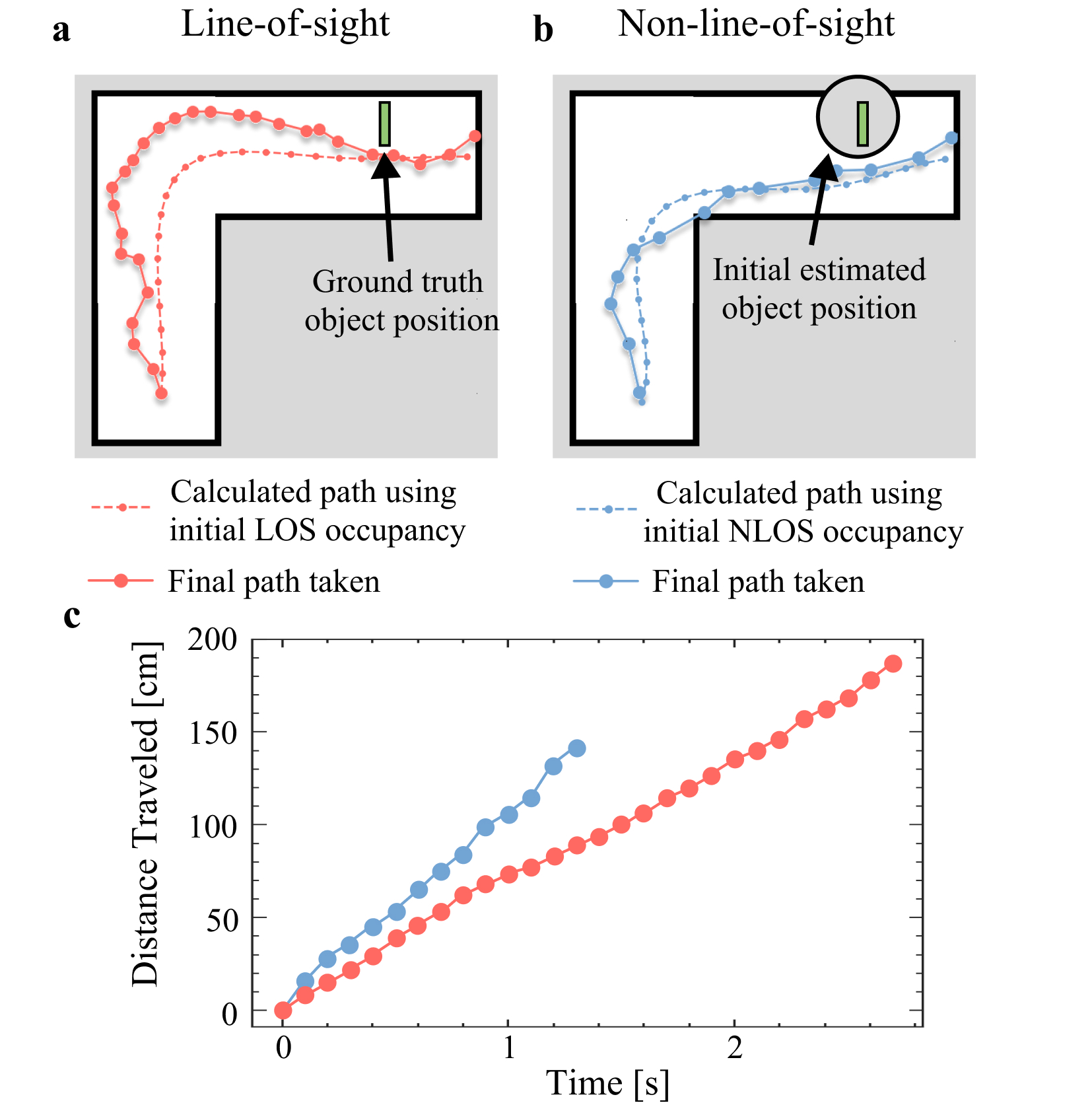}
    \caption{\textbf{Blind corner navigation using non-line-of-sight perception on a physical robot.} To test the effectiveness of non-line-of-sight imaging for autonomous navigation, we ran experiments on a physical robot in an L-shaped blind corner. An object was placed in an area hidden from view of the sensor. (a) The agent with only a LOS view couldn't prepare for the object before rounding the corner, and had to overshoot back inside to reach the goal. (b) On the other hand, when NLOS information was used, the robot could successfully generate and follow a path which reduced the wasted travel time. (c) A quantitative depiction of the efficiency improvements for an NLOS-capable robot. With only LOS information, the robot took more than 2$\times$ longer and required a 33\% longer trajectory.}
    \label{fig:real-results}
\end{figure}

To demonstrate the usefulness of third-bounce light for robotic navigation, we deploy a mobile robot in an L-shaped corridor. We then compare its navigation performance using strictly line-of-sight information versus non-line-of-sight measurements. As shown in \Cref{fig:real-results}a and \Cref{fig:real-results}b, without information about the hidden object (i.e. LOS only), the algorithm essentially produces a centerline path through the corridor. This approach is suboptimal, as the robot encounters the object as soon as it rounds the corner, thus needing costly maneuvers later on. Ideally, the agent should anticipate the obstacle and take a sharper path closer to the inside of the corner. In the non-line-of-sight case (\Cref{fig:real-results}b), the robot localizes the hidden object from the outset, enabling it to plan an efficient path that immediately avoids the obstacle.

By effectively seeing around the corner, the agent is able to produce more desireable control outcomes. As shown in \Cref{fig:real-results}c, by planning a route that avoids sharp maneuvers later in the experiment, the NLOS-enabled agent follows a route ~33\% shorter than if it were using strictly LOS information. Furthermore, by following a longer route, the robot that uses LOS information takes nearly 2$\times$ the amount of time than in the NLOS case.
\section{Conclusion and Future Work}
\label{sec:conclusion}
In this paper, we demonstrate the utility of NLOS perception using single-photon LiDAR for autonomous navigation. Our novel pipeline estimates the occupancy map of the hidden region from multi-bounce transients and then utilizes the occupancy map for autonomous navigation. In simulations, we show that our approach, with NLOS perception, reduces the likelihood of collisions with the hidden target at high speeds compared to traditional LOS-only perception. Through experiments with a commercial-grade single-photon LiDAR mounted on a two-wheeled robot, we demonstrate a smoother trajectory in navigating around hidden objects. 

Our work opens up new avenues for sensing, perception, and control systems that leverage indirect reflections for safer navigation around hidden obstacles. There are several areas for future improvement and investigation. On the sensing front, addressing the low signal-to-noise ratio of commercial-grade single-photon LiDARs in measuring multi-bounce light remains challenging. While our initial evaluation utilized retro-reflective surfaces, developing robust approaches for diffuse surfaces is a crucial next step. Regarding perception, our data-driven approach could be extended by exploring the use of dynamics-integrated transient renderers for training on simulated data, potentially reducing the need for extensive physical data collection. Finally, there is significant potential to enhance our control framework to handle more complex environments and multiple hidden targets, further improving the versatility and applicability of NLOS perception in autonomous navigation.
\section{Acknowledgments}
\label{sec:acknowledge}
This work was supported in part through NSF project CMMI2153855. Aaron Young was supported by the NSF GRFP (No.
2022339767).


\bibliographystyle{unsrt}


\addtolength{\textheight}{-12cm}   


\end{document}